\documentclass[journal]{IEEEtran}
\ifCLASSINFOpdf
\else
\fi
\usepackage{graphicx}
\usepackage{amsmath}
\usepackage{cite}
\usepackage{booktabs}
\usepackage{multirow}
\usepackage{bibentry}
\usepackage{svg}
\usepackage{amssymb}
\usepackage{float}

\usepackage[linesnumbered,ruled]{algorithm2e}
\usepackage{algpseudocode}

\usepackage[numbers,sort&compress]{natbib}
\begin{document}
%
\title{SgVA-CLIP: Semantic-guided Visual Adapting of Vision-Language Models for Few-shot Image Classification}
%
%
%

\author{{Fang~Peng,
        Xiaoshan~Yang, Linhui~Xiao, Yaowei~Wang and~Changsheng~Xu, \IEEEmembership{Fellow,~IEEE}}

\thanks{Fang Peng and Linhui Xiao are with the National Laboratory of Pattern Recognition, Institute of Automation, Chinese Academy of Sciences, Beijing 100190, China, also with the Peng Cheng Laboratory, Shenzhen 518066, China, and also with the School of Artificial Intelligence, University of Chinese Academy of Sciences, Beijing 100049, China (e-mail: pengfang21@mails.ucas.ac.cn, xiaolinhui16@mails.ucas.ac.cn).}
\thanks{Yaowei Wang is with the Peng Cheng Laboratory, Shenzhen 518066, China (e-mail: wangyw@pcl.ac.cn).}
\thanks{Xiaoshan Yang and Changsheng Xu are with the National Laboratory of Pattern Recognition,
Institute of Automation, Chinese Academy of Sciences, Beijing 100190,
China, also with the School of Artificial Intelligence, University of Chinese Academy of Sciences, Beijing 100049, China, and also with the Peng Cheng Laboratory, Shenzhen 518066, China (e-mail: xiaoshan.yang@nlpr.ia.ac.cn, csxu@nlpr.ia.ac.cn).}
\thanks{Changsheng Xu is the corresponding author.}}


%

\markboth{Journal of \LaTeX\ Class Files,~Vol.~14, No.~28, November~2022}
{Shell \MakeLowercase{\textit{et al.}}: Bare Demo of IEEEtran.cls for IEEE Journals}
%



\maketitle

\begin{abstract}
Although significant progress has been made in few-shot learning, most of existing few-shot image classification methods require supervised pre-training on a large amount of samples of base classes, which limits their generalization ability in real world application. Recently, large-scale Vision-Language Pre-trained models (VLPs) have been gaining increasing attention in few-shot learning because they can provide a new paradigm for transferable visual representation learning with easily available text on the Web. However, the VLPs may neglect detailed visual information that is difficult to describe by language sentences, but important for learning an effective classifier to distinguish different images. To address the above problem, we propose a new framework, named Semantic-guided Visual Adapting (SgVA), which can effectively extend vision-language pre-trained models to produce discriminative adapted visual features by comprehensively using an implicit knowledge distillation, a vision-specific contrastive loss, and a cross-modal contrastive loss. The implicit knowledge distillation is designed to transfer the fine-grained cross-modal knowledge to guide the updating of the vision adapter. State-of-the-art results on 13 datasets demonstrate that the adapted visual features can well complement the cross-modal features to improve few-shot image classification.
\end{abstract}


%
\IEEEpeerreviewmaketitle

\begin{figure}[t]
\centering
\includegraphics[width=1.0\linewidth]{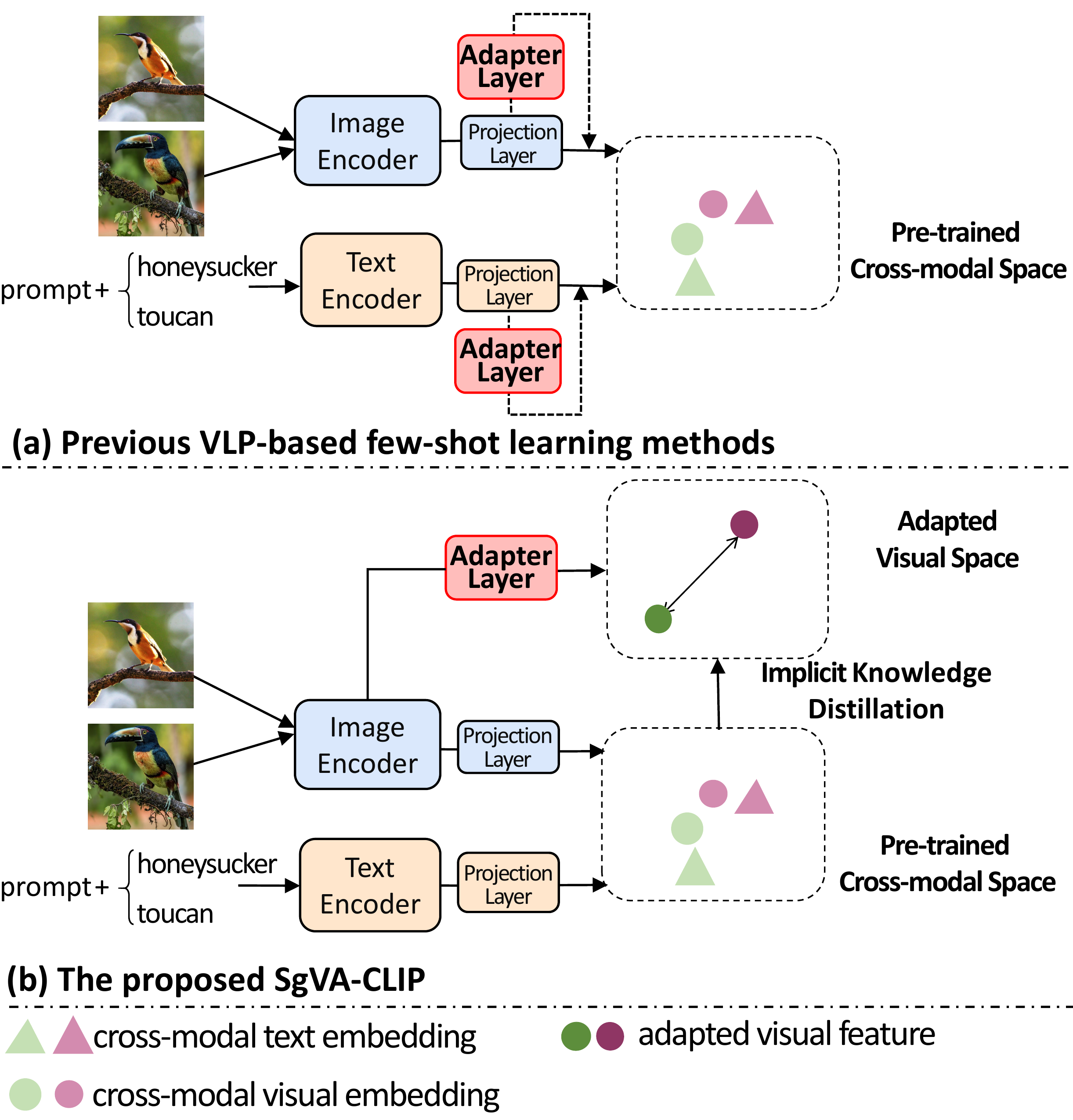}
\caption{
SgVA-CLIP vs. previous VLP-based few-shot learning methods.
(a) Previous VLP-based few-shot learning methods focus on enhancing the cross-modal alignment, which may neglect important task-specific visual information (e.g., the two birds \textit{honeysucker} and \textit{toucan} have different beaks) for distinguishing different images when the labeled samples are insufficient.
%
%
(b) SgVA-CLIP makes a comprehensive consideration of adapted visual feature space and pre-trained cross-modal feature space.
%
The adapted visual features provide more discriminative visual information and thus can well complement the cross-modal features to improve few-shot image classification.}
\label{intro}
\end{figure}
%

\section{Introduction}
Few-shot learning refers to the task of learning a new concept with only a few labeled samples,
which is inspired by human learning ability.
As labeling is often
expensive in real scenarios, few-shot learning has become an important and widely studied problem.
%
%
However, with little supervision information, learning to recognize new classes is challenging because directly training the model from a few samples may overfit.
A common idea in few-shot learning is to train the model from base classes with sufficient samples to get prior knowledge and then migrate to the novel classes with a few examples.
%
%
%
Existing studies on few-shot image classification can be roughly divided into three categories, namely fine-tuning based methods, data augmentation based methods and meta learning based methods.
Among them, the most widely studied is meta-learning \cite{hospedales2021meta,9678031,9665225}, which acquires the abstract learning ability to generalize to new classes by learning meta-knowledge from a set of different meta tasks.
Although significant progress has been made in few-shot learning,
most of existing few-shot learning methods require the network to be pre-trained in a supervised manner on a large amount of labeled data of base classes.
%
%
As a result, the current few-shot learning methods have limited generalization ability and is impractical in the real world 
due to the shortage of supervised data.


%
%
Recently, self-supervised learning \cite{chen2020simple,he2022masked,9811387,9714824} has emerged as a possible solution to alleviate the dependency on large-scale labeled data.
%
%
Self-supervised learning exploits pretext tasks to mine supervised information from large-scale unsupervised data, thereby learning rich implicit priors and latent representations.
%
%
%
With the development of self-supervised learning,
Vision-Language Pre-trained models (VLPs),
e.g., CLIP \cite{radford2021learning}, ALIGN \cite{jia2021scaling}, and Florence \cite{yuan2021florence}, attract more and more attention due to its significant performance in a variety of downstream tasks, such as image classification and visual question answering.
%
%
%
%
%
%
VLPs like CLIP can provide effective visual and semantic knowledge of open-world concepts that are learned on large-scale image-text pairs, laying a good generalization foundation of few-shot image classification.\par
VLPs have been successfully applied to few-shot image classification 
with the help of carefully designed text prompts \cite{radford2021learning,9914670}, which can change the discrete class labels into language sentences.
%
%
For example, CLIP model \cite{radford2021learning} learns vision and language representations by aligning the image and text in a cross-modal joint space, which allows images to be correctly classified via image-text similarity.
%
%
There are also VLPs-based few-shot learning methods that focus on enhancing the image-text alignment.
%
%
Context Optimization (CoOp) \cite{zhou2022learning} is proposed to improve the text embedding of CLIP by soft prompt engineering.
%
CLIP-Adapter \cite{gao2021clip} fine-tunes the image representation by adjusting an extra bottleneck layer.
{
ProGrad \cite{BeierZhu2022PromptalignedGF} proposes Prompt-aligned Gradient to prevent prompt tuning from forgetting the general
knowledge learned from VLPs.}
%

%
Existing methods only consider the image-text alignment when transferring the VLPs to solve few-shot image classification.
%
%
Although relying on the image-text similarity can well capture the visual and semantic knowledge learned by the pre-trained model,
it is sometimes unreliable to recognize objects without comprehensively considering the specific discriminative visual information of the few-shot task.
%
%
%
The reason is that to learn a good image-text alignment model, the pre-trained VLPs may neglect detailed visual information that is difficult to describe by language sentences.
However, the neglected visual information is probably important for distinguishing different images when the labeled samples are insufficient.
%
%
%
%
%

%
To address the above problem, we propose a new framework, named Semantic-guided Visual Adapting (SgVA), which can effectively extend vision-language
pre-trained models (e.g., CLIP) to produce discriminative adapted visual features with the guidance of the fine-grained cross-modal knowledge learned by the pre-trained model.
The adapted visual features can well complement the cross-modal features to improve few-shot image classification.
Fig. \ref{intro} shows the main idea of our work.
Specifically,
our method is extended from the pre-trained CLIP.
Given labeled support images and unlabeled query images, we firstly extract the visual features for the images from the output before the cross-modal projection layer of the CLIP model, which are referred to as pre-trained visual features. And cross-modal embeddings for both the images and the prompted texts of the class labels are extracted from the output of the cross-modal projection layer.
Next, we map the pre-trained visual features to the adapted visual features by a visual adapting layer.
We update the visual adapting layer on the few-shot samples by a vision-specific contrastive loss and cross-modal contrastive loss
with the help of vision prototypes and cross-modal prototypes that are obtained by averaging the corresponding sample features of a given class.
Moreover, we adopt an implicit distillation to utilize the fine-grained cross-modal knowledge (i.e., relative similarities between samples and prototypes in the {pre-trained} cross-modal space) learned by the pre-trained model to guide the updating of the vision adapter.
%
%
%
Finally, we infer the class label for a given query sample by jointly considering its distance to vision-specific prototypes and cross-modal prototypes.
%
%
%
%
%
\par

%
%

%
Our contributions are summarized as follows.
%
%
We propose a new framework of semantic-guided visual adapting, 
which flexibly extends the vision-language pre-trained models (e.g., CLIP) to produce discriminative adapted visual features by jointly using implicit knowledge distillation, vision-specific contrastive loss, and cross-modal contrastive loss.
%
%
%
%
%
%
%
%
We obtain new state-of-the-art results in few-shot image classification by comprehensively considering the sample relations based on both the adapted visual features and the cross-modal features,
which demonstrates a strong complementarity between the two kinds of feature space and also provides a promising direction for future research.

\section{Related Work}
This section reviews three topics closely related to our work in terms of few-shot learning, prototype networks and knowledge distillation.

\subsection{Few-shot Learning}
%
%
Few-shot learning aims to learn a model that can recognize new classes with a few training samples.
The widely studied conventional few-shot learning methods include {fine-tuning \cite{vedaldi2014convolutional,shen2021partial}, data/feature augmentation~\cite{kumar2019closer,yoon2019tapnet}, 
and meta learning \cite{hospedales2021meta,sun2019meta}}.
%
Recently, 
%
%
Vision-language pre-trained models (VLPs) (e.g. CLIP \cite{radford2021learning} and ALIGN \cite{jia2021scaling}) have been applied to few-shot learning by transferring the powerful representation ability.
%
%
%
%
%
%
%
%
In order to realize data-efficient fine-tuning, CoOp \cite{zhou2022learning} improves the ability of image-text alignment through continuous prompt optimization,
and CLIP-Adapter \cite{gao2021clip} designs lightweight feature adapters to explore simple fine-tuning.
After that, Tip-Adapter \cite{zhang2021tip}, a training-free method, is proposed to save computational resources.
Different from them, VT-CLIP \cite{zhang2021vt} improves the interaction of image and text branches of CLIP by cross-modal module.
Other works like MUST \cite{li2022masked} and UPL \cite{huang2022unsupervised} think about unsupervised learning.
Besides, WiSE-FT \cite{wortsman2022robust} and CoCoOp \cite{zhou2022conditional} consider both the accuracy of target distribution and robustness to distribution shifts.
Unlike the above methods that focus on image-text contrastive learning, we extend the 
pre-trained CLIP to learn more discriminative visual features that can well complement the cross-modal features in few-shot learning.
%
%
 
\subsection{Prototype Networks}
%
Prototype network \cite{snell2017prototypical} is proposed in 2017 to solve the problem of few-shot classification,
%
which aims at learning a metric space where query samples can be accurately classified by calculating the distances between queries and prototypes.
Compared with other few-shot learning methods, Prototype network reflects a simpler inductive bias, which is beneficial in the case of limited data.
Owing to the potential of this paradigm, many variations have been developed since then.
%
%
Chen \emph{et al}. \cite{chen2020new} found that introducing an extra pre-training phase on the entire base classes could improve performance, but it leads to poor generalization ability.
%
%
Early prototype networks only employ visual information, but increasingly there are approaches to explore how semantic knowledge can enhance the performance.
For example, Chen \emph{et al}. \cite{xing2019adaptive} learned semantic knowledge from unsupervised corpora, and proposed an adaptive modality mixing mechanism to combine the visual and semantics knowledge, showing improvements in few-shot learning.
Frederik \emph{et al}. \cite{pahde2021multimodal} mapped text data to the visual embedding space with the help of a generative model, and then designed a strategy to combine the real and generated features through the nearest neighbor algorithm.
%
%
Instead of only forming a single metric space as in existing prototype networks, we construct two metric spaces including visual and cross-modal spaces to comprehensively conduct the few-shot learning.
%



\subsection{Knowledge Distillation}
%
Knowledge Distillation (KD) \cite{bucilua2006model,hinton2015distilling,10015053} means transferring the knowledge from the pre-trained complex model (teacher model) to a simpler structured network (student model).
Owing to its superior performance in knowledge transferring and model enhancement, KD is widely used in model compression and transfer learning.
In the process of KD, the output of teacher model is used as the supervision signal to train the student model through the distillation loss.
And the optimization target is to make the class-level probability distribution of the student model match the probability output of the teacher model.
In terms of model compression, 
DistillBert \cite{sanh2019distilbert} and TinyBert \cite{jiao2019tinybert} use KD to explore smaller and faster models for language representation learning.
%
%
In addition to model compression, KD also plays an important role in knowledge transferring between different modalities.
For example,  the vision-language distillation framework DistillVLM \cite{fang2021compressing} is proposed to improve vision-language tasks like image captioning and VQA.
Hafner \emph{et al}. \cite{hafner2022cross} propose a novel cross-modal distillation method for robust person re-identification, which transfers knowledge from RGB images to depth images.
Besides, in \cite{cho2020speech}, semantic knowledge is transmitted from language model to a spoken language understanding module, so that the deficiency of speech data can be alleviated.
The above works show the benefits of KD in both self-supervised learning and multi-modal alignment.
%
Different from existing methods, we propose a knowledge distillation which can implicitly transfer the fine-grained relation knowledge learned in the cross-modal space to the visual space by conducting the distillation in the cross-modal space.
%
%
%
%

\par

\section{The Proposed Method}
\subsection{Problem Definition}
{
The {few-shot learning methods} usually depend on base classes with ample samples to learn {how to generalize to novel classes}. However, in the real few-shot scenario, sufficient samples of base classes may not be available. It is worth noting that since the pre-trained vision and language model, e.g., CLIP, can provide a good foundation model for the few-shot learning, the base classes are not indispensable.
In this work, we consider both the standard meta-learning scenario with base classes and the scenario without base classes.
%

%
%
%
In the standard meta-learning scenario, a series of meta-tasks (episodes) are created for training and testing.
For each meta-task in the {meta-training} phase, $N$ (Way) base classes and $K$ (Shot) samples for each base class are randomly selected to make up the Support Set $S \!= \!\left \{ \left ( \boldsymbol{x_{i}}, y_{i}   \right )  \right \}_{i = 1}^{N\times K}$, where $\boldsymbol{x_{i}}$ denotes the sampled image, and $y_{i}$ is the label of $\boldsymbol{x_{i}}$.
Other $M$ samples of the $N$ (Way) classes
are randomly selected to form the Query Set $Q = \left \{ \left ( \boldsymbol{x_{i}}, y_{i}   \right )  \right \}_{i = 1}^{M}$.
In the {meta-test phase}, the Support Set is created from $K$-shot labeled samples of novel classes and the query set is created from unlabeled samples of novel classes.
%
%
%



%
In the scenario without base classes, inspired by the idea of self-supporting from \cite{fan2022self}, we build both the Support Set and the Query Set from the $K$-shot labeled samples of novel classes in the meta-training phase.
%
%
In the {meta-test phase}, 
%
the Support Set is created from $K$-shot labeled samples of novel classes and the Query Set is created from unlabeled samples of novel classes.
%
%
%
%

%

\subsection{Network Architecture}\label{subsec:framework}

\begin{figure*}[t]
\centering
\includegraphics[width=1.0\linewidth]{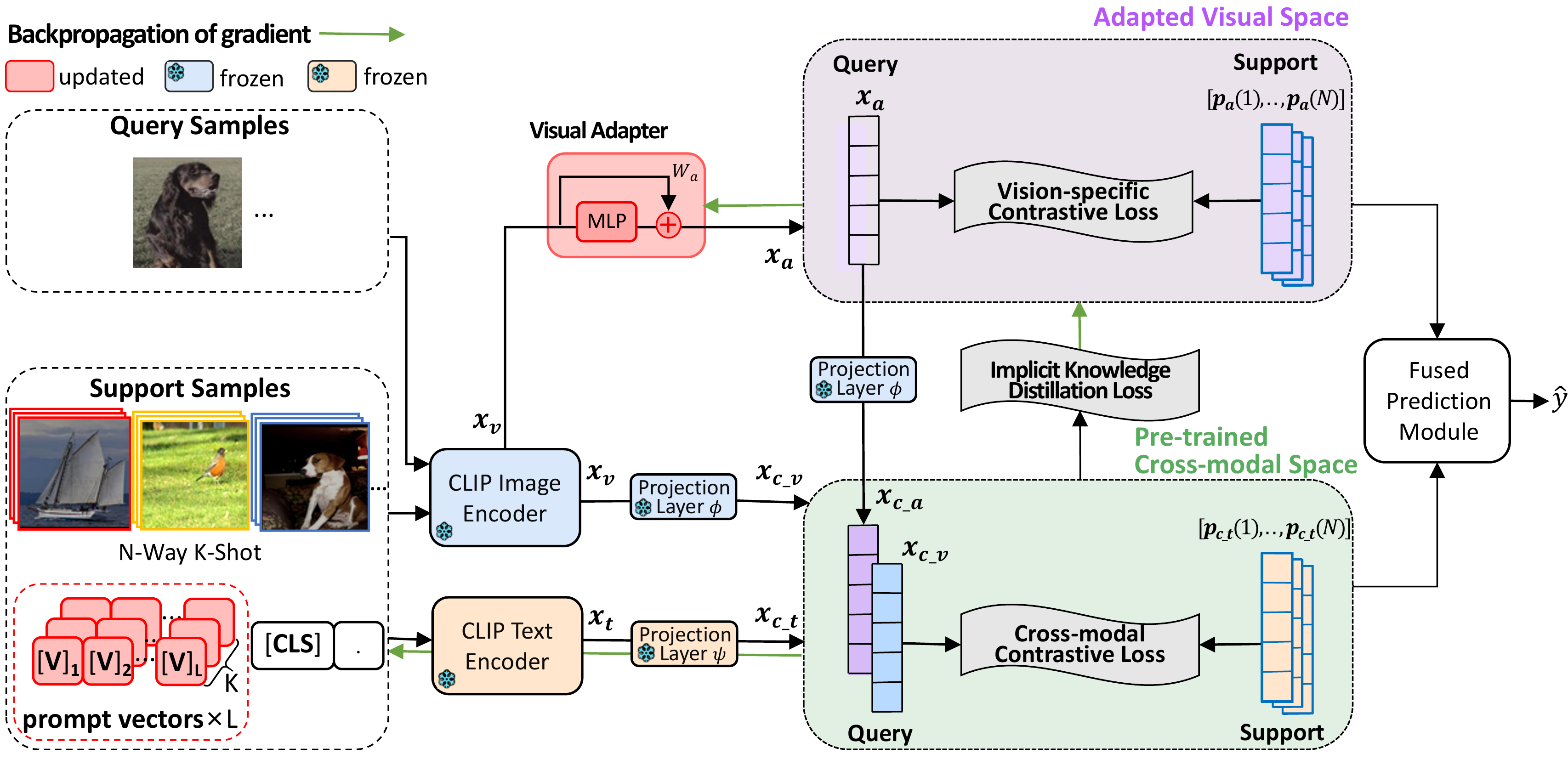}
\caption{
Overview of the proposed SgVA-CLIP.
%
%
In the pre-trained cross-modal space, 
we build cross-modal prototypes that are obtained by averaging the cross-modal text embedding, i.e., $\boldsymbol {x_{c\_t}}$ of the support samples.
Meanwhile, in the adapted visual space, we build vision-specific prototypes that are obtained by averaging the adapted visual features after the visual adapting layer, i.e., $\boldsymbol {x_{a}}$ of the corresponding support samples.
The SgVA-CLIP can learn discriminative adapted visual features with the guidance of the cross-modal knowledge learned by the pre-trained CLIP.
And the discriminative visual features can well complement the cross-modal features in the few-shot image classification.}
\label{model_overview}
\end{figure*}

%
%

The proposed framework, namely Semantic-guided Visual Adapting (SgVA), aims to learn discriminative adapted visual features with the guidance of the cross-modal knowledge learned by the pre-trained CLIP model.
The adapted visual features can well complement the cross-modal features in the few-shot image classification.
%
%
Fig. \ref{model_overview} shows the overall framework.
%
Given support images and query images in each episode,
%
%
%
the class labels of support samples are firstly transformed into text input by L-length learnable prompt vectors as in \cite{zhou2022learning}.
%
Then, the images and texts go through the frozen image and text encoders of CLIP 
to generate the pre-trained visual features ($\boldsymbol {x_v}$) and text features ($\boldsymbol {x_t}$) respectively, which are further mapped to cross-modal visual and text embeddings, i.e., $\boldsymbol {x_{c\_v}}$ and $\boldsymbol {x_{c\_t}}$ 
through different projection layers (i.e., $\phi$ and $\psi$).
%
%
%
%
Meanwhile, 
the pre-trained visual features ($\boldsymbol {x_v}$) are mapped to the adapted visual features ($\boldsymbol {x_a}$) by a visual adapting layer.
To fully exploit sample relations in the visual space and cross-modal space,
we build vision-specific prototypes and cross-modal prototypes that are obtained by averaging the corresponding support sample features of a given class.
Based on the prototypes, we update the visual adapting layer on the few-shot samples by an implicit knowledge distillation, a vision-specific contrastive loss, and a cross-modal contrastive loss.
%
%
%
The implicit knowledge distillation utilizes fine-grained sample relations in the cross-modal feature space 
to guide the learning of the visual adapting to produce more discriminative visual features.
%
In the inference process, test image can be recognized by jointly considering its distance to the vision-specific prototypes and cross-modal prototypes.
%


%

\subsection{Shot-specific Text Prompt}
{
To transform the class label into natural language sentence that can be directly processed by CLIP, we adopt prompted text with individual differences as in~\cite{zhang2022prompting}.
We independently initialize an individual prompt $\boldsymbol t_i$ for 
each shot from the same class.
We use $K$ different prompts in each $N$-way $K$-shot episode according to the number of shots and we use shared prompts for different classes, which is referred to as shot-specific text prompt.} 
%
As illustrated in~\cite{zhang2022prompting}, using a fixed amount of prompts instead of a universal
prompt can capture subtle differences among different samples and can also be more efficient than designing a specific prompt for every input sample.
The text prompt $\boldsymbol t_i$ is formally defined as the concatenation of $L$ learnable continuous vectors and the class name embedding:
\begin{equation}\boldsymbol t_i = [\boldsymbol V]_{i,1}[\boldsymbol V]_{i,2} . . . [\boldsymbol V]_{i,L}[\mathrm{\boldsymbol {CLS}}]_i[.],\label{text_form}\end{equation} where each $[\boldsymbol V]_{i,l}$, $l \in \{1, . . . , L\}$, is a learnable vector with the same dimension as the class embedding $[\mathrm{\boldsymbol {CLS}}]_i$.
%
The class embedding is a 512-dimensional word embedding obtained from 
the pre-trained CLIP.

\subsection{Visual Adapting Layer}
%
%
{\color{red}
}

%
%
The adapting layer consists of a two-layer Multi-layer Perceptron (MLP) and an adaptive residual connection.
The new feature $New_v(\boldsymbol{x_v})$ acquired in the adapting layer can be represented as:
%
\begin{equation}
New_{v}(\boldsymbol{x_v}) =  \mathrm{ReLU}(\boldsymbol{x_v}^{T} \boldsymbol {W_1})\boldsymbol {W_2}, \label{new_feature} \end{equation}

where $\boldsymbol{x_v}$ denotes the pre-trained visual feature {that is obtained by the image encoder of the pre-trained CLIP encoding the image of the sample $\boldsymbol x$ before the linear projection to the cross-modal embedding space}, $\boldsymbol{W_1}$ and $\boldsymbol{W_2}$ are learnable weights of the two fully connected layers, and ReLU is activation function. The pre-trained visual feature $\boldsymbol{x_v}$ is fixed in our framework.
%
Then, the new feature is added to the original visual feature by an adaptive residual connection:
\begin{equation} \boldsymbol{x_a} =  [New_{v}(\boldsymbol{x_v}), \boldsymbol{x_v}]\boldsymbol{W_a}, \label{xa_form}\end{equation}
where $\boldsymbol{x_a}$ is the adapted visual feature, and the $\boldsymbol{W_a}\in \mathbb{R}^{2}$ is a learnable weight vector. $[]$ denotes the operation of combining the vectors as a matrix.

\subsection{Cross-modal Contrastive Loss}
%
%
%
%
%
%

For the convenience of exploring the relations between different classes in the cross-modal space, 
we firstly calculate cross-modal prototypes ($\boldsymbol {p_{c\_t}}$) of $N$ classes by averaging the cross-modal text embeddings of the support samples for each class.
%
The cross-modal prototype of the $k$-th class, i.e., $\boldsymbol {p_{c\_t}}(k)$ is represented by \begin{equation}
\mathbf{\boldsymbol {p_{c\_t}}}(k)=\frac{1}{\left|S(k)\right|} \sum_{\left({\boldsymbol{x_{i}}}, y_{i}\right) \in S(k)} \boldsymbol {x_{c\_t}}(i),
\label{cross-modal prototype}
\end{equation}where $\boldsymbol {x_{c\_t}}(i)$ denotes the cross-modal text embedding of the $i$-th sample that belongs to $S(k)$, and $S(k)$ is the set of all the support samples of class $k$.
As illustrated in Section~\ref{subsec:framework},
$\boldsymbol{x_{c\_t}}$ is obtained by the pre-trained CLIP that is fixed in our framework.
The prompted text ($\boldsymbol t_i$) of the $i$-th sample is encoded by the text encoder of CLIP into the uni-modal text feature $\boldsymbol {x_{t}}(i)$, which is then projected into the cross-modal embedding space as cross-modal text embedding $\boldsymbol {x_{c\_t}}(i)$.

The purpose of cross-modal contrast loss is to learn a better textual representation and lay a good semantic foundation for guiding the visual adapting.
%
%
For every query sample $\boldsymbol x$, the cosine similarity scores between the cross-modal visual feature  ($\boldsymbol {x_{c\_v}}$) and the cross-modal prototypes ($\boldsymbol {p_{c\_t}}$) are calculated, where $\boldsymbol {x_{c\_v}}$ is obtained by mapping the pre-trained visual feature into the cross-modal embedding space.
The cosine similarity is scaled by a temperature parameter $\tau_1$, and normalized into a probability distribution via Softmax.
Then, the cross-modal contrastive loss $L_{cl \_i2t}$ is defined as the cross entropy loss over the probability distribution:
\begin{equation}
L_{cl \_i2t}=-log\frac{exp (< \boldsymbol{x_{c\_v}}, \boldsymbol{{p_{c\_t}}_{(+)}} > /\tau_{1})}{ { \sum_{j=1}^{N}} exp (<\boldsymbol{{x_{c\_v}}}, \boldsymbol{p_{c\_t}}(j) > /\tau_{1})},
\label{i2t}
\end{equation}
where $\tau_{1}$ is a temperature parameter that is initialized to 0.07 and pre-trained by CLIP \cite{radford2021learning}. 
%
And $\boldsymbol{{p_{c\_t}}_{(+)}} $ denotes the the cross-modal prototype of the positive class,
%
$< \cdot , \cdot >$ denotes cosine similarity, and $N$ is the number of classes.
The useful implicit knowledge is included in the relative distances between cross-modal features.

\subsection{Vision-specific Contrastive Loss}
%
%
Since the pre-trained CLIP may neglect important visual information that is difficult to describe in natural language sentences,
we utilize vision-specific contrastive loss to make the adapted features (i.e., $\boldsymbol{x_a}$) retain more discriminative visual information that is specific to the current few-shot task.
%
%
%
For every query sample $\boldsymbol x$, the cosine similarity scores between the adapted visual feature ($\boldsymbol {x_{a}}$) and the vision-specific prototypes ($\boldsymbol {p_{a}}$) of $N$ classes are calculated by: \begin{equation}
\mathbf{\boldsymbol{p}_{a}}(k)=\frac{1}{\left|S(k)\right|} \sum_{\left({\boldsymbol{x_{i}}}, y_{i}\right) \in S(k)} \boldsymbol {x_{a}}(i),
\label{vision prototype}
\end{equation}where $S(k)$ denotes the set of all the support samples of class $k$ and {$\boldsymbol{x_a}(i)$ is calculated by Eq. \ref{new_feature} and Eq. \ref{xa_form}} for the $i$-th sample.
%
%
The purpose of vision-specific contrastive loss is to maximize the cosine similarity between $\boldsymbol{x_{a}}$ and 
the positive prototype
while minimizing the cosine similarity between $\boldsymbol {x_{a}}$ and negative prototypes.
%
%
Formally, the vision-specific contrastive loss $L_{cl \_i2i}$ is calculated by: 
\begin{equation}
L_{cl \_i2i}=-log\frac{exp (< \boldsymbol{x_{a}}, \boldsymbol{{p_{a}}_{(+)}} > /\tau_{1})}{ { \sum_{j=1}^{N}} exp (<\boldsymbol{{x_{a}}}, \boldsymbol{p_{a}}(j) > /\tau_{1})}.
\label{i2i}
\end{equation}

\subsection{Implicit Knowledge Distillation}

\begin{figure}[t]
\centering
\includegraphics[width=0.5\textwidth]{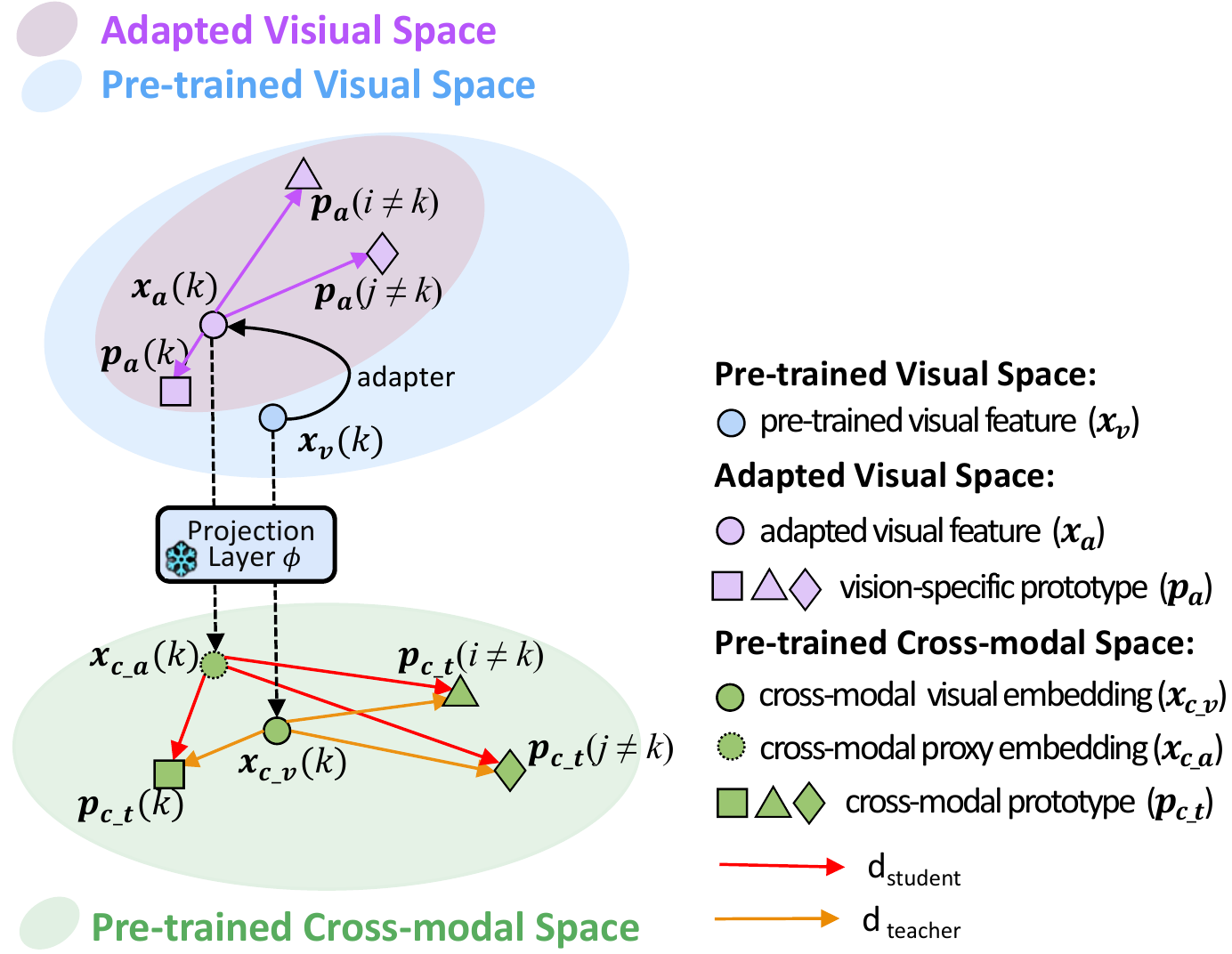}
\caption{Interpretation of the implicit knowledge distillation between the adapted visual space and the {pre-trained} cross-modal space. The adapted visual feature $\boldsymbol{x_{a}}$ is mapped to the cross-modal space as a proxy representation $\boldsymbol{x_{c\_a}}$. The knowledge distillation is performed in the {pre-trained} cross-modal space by matching the fine-grained sample-prototype relations {for} the proxy representation $\boldsymbol{x_{c\_a}}$ and the cross-modal visual embedding $\boldsymbol{x_{c\_v}}$.}
\label{fig2}
\end{figure}

With the vision-specific contrastive loss,
we can already make the samples of the same class close to each other, and the samples of different classes far from each other in the adapted unimodal vision space.
However, the vision contrastive loss cannot provide the fine-grained relations between different samples.
%
As shown in Fig.~\ref{intro}, the two kinds of birds have different beaks locally, but the whole is similar in vision. If the vision-specific contrastive loss is used alone, it is easy to ignore the details of beaks, and lead to misclassifying the bird. However, they are easy to distinguish in terms of text semantics (i.e., honeysucker and toucan). The cross-modal information in the pre-trained cross-modal space provides the constraint of relative semantic distances between samples, and instructs how far apart the semantically different samples should be, even if they are visually similar as a whole, thus helping to improve the discriminative ability of the adapted visual features.

%
%
%
As a solution,
we consider to take advantage of the relative sample relationships produced by the pre-trained CLIP model, which cannot be achieved by contrastive learning alone.
%
%
It is worth noting that when compared with the visual data, the semantic meaning of a given class is more likely to be shared by the pre-trained CLIP model and the downstream few-shot task.
%
Therefore, we adopt a knowledge distillation loss 
to guide the learning of the vision adapter
by the fine-grained relationships in the cross-modal space.
%
%

%
%
Fig. \ref{fig2} gives an illustration of the proposed implicit knowledge distillation.
%
By mapping the adapted visual feature $\boldsymbol {x_{a}}$ to the cross-modal space through the projection layer (i.e., $\phi$), a proxy representation of it in the cross-modal space, namely $\boldsymbol {x_{c\_a}}$ is obtained.
%
Then, we can utilize the fine-grained relations of the cross-modal visual feature $\boldsymbol {x_{c\_v}}$ to constrain the proxy representation $\boldsymbol {x_{c\_a}}$.
This means that we can use distances between samples obtained in the cross-modal space as extra supervision to implicitly guide the learning of the vision adapting layer. 
The knowledge distillation is performed in the cross-modal space, and the gradients are backpropagated to the visual adapting layer.
%
%

%

%
More specifically, for a cross-modal visual feature $\boldsymbol {x_{c\_v}}$, 
we calculate its distances to each of the cross-modal text prototypes as a teacher:\begin{equation}
\boldsymbol {d_{tea}}(k)=< \boldsymbol{x_{c\_v}}, \boldsymbol{{p_{c\_t}}}(k) > /\tau_{1},
\label{logits_i2t}
\end{equation}
where $\tau_{1}$ {is a temperature parameter that is initialized to 0.07, and learned by the pre-trained CLIP.}
%
%
The distances between $\boldsymbol {x_{c\_a}}$ and each of the cross-modal text prototypes $\boldsymbol{p_{c\_t}}$ 
are calculated as a student:
\begin{equation}
\boldsymbol {d_{stu}}(k)=< \boldsymbol{x_{c\_a}}, \boldsymbol{{p_{c\_t}}}(k) > /\tau_{1}.
\label{logits_newi2t}
\end{equation}
%
%
Finally, the knowledge distillation loss is defined as:\begin{equation}
\begin{small}
\begin{aligned}
L_{KD} \!=\! - \! \sum_{k=1}^{N} 
 \frac{exp(\boldsymbol {d_{tea}}(k)/\tau_{2})}{ \sum_{j=1}^{N} exp(\boldsymbol {d_{tea}}(j)/\tau_{2})} log \Big(\frac{exp(\boldsymbol {d_{stu}}(k)/\tau_{2})}{ \sum_{j=1}^{N}  exp(\boldsymbol {d_{stu}}(j)/\tau_{2})} \Big),
\end{aligned}
\end{small}
\label{loss_kd}
\end{equation}
where $\tau_{2}$ is a hyperparameter that is set to 5 in this paper. The parameter analysis of $\tau_{2}$ is shown in Table \ref{table5}.
%

\subsection{Optimization and Inference}
%
Our framework can be learned in an end-to-end form for few-shot image classification.
The framework is optimized by jointly considering vision-specific contrastive loss, cross-modal contrastive loss, and knowledge distillation loss.
%
%
%
%
The overall loss is defined as:
\begin{equation}
Loss = L_{cl \_i2t} + L_{cl \_i2i} + L_{KD}.
\label{overall_loss}
\end{equation}
%
%
}

%
%
%

For a given test image in the inference phase, we conduct few-shot classification by comprehensively considering its distance to the vision-specific prototypes and cross-modal prototypes:
{
\begin{align}
\boldsymbol{d_{a}}(k) = <\boldsymbol{x_{a}},  \boldsymbol {p_{a}}(k)>, \quad \quad \quad  k=1,...,N,\\
\boldsymbol{d_{c\_t}}(k)= <\boldsymbol{x_{c\_v}},  \boldsymbol {p_{c\_t}}(k)>, \ \ \ k=1,...,N,
\end{align}
where $< \cdot, \cdot >$ denotes cosine similarity.
%
%
%
%
%
Specifically, 
the predicted class $\hat{y}$ with the maximum posterior probability is calculated by applying the Naive Bayes:
\begin{equation}
\hat{y}=\mathop{\mathrm{argmax}}\limits_{y_k}\ \! {{ \sum_{j=1}^{2N}} \log_{}{P\big(\boldsymbol {d}{(j)} \mid Y\!=\!y_k\big)}} ,
\label{yhat}
\end{equation}
where $\boldsymbol {d}$ is a $2N$-dimensional vector obtained by concatenating $\boldsymbol{d_{a}}$ and $\boldsymbol{d_{c\_t}}$.
%
%
%
%
}

\begin{table*}[t]
\centering
\caption{Comparison with state-of-the-art few-shot learning methods (with base classes) on miniImagenet and tieredImagenet.}
\setlength{\tabcolsep}{2.80mm}
\renewcommand\arraystretch{1.3}
\begin{tabular}{l|c|cc|cc}
\toprule[1pt]
 &  & \multicolumn{2}{c|}{miniImagenet (\%)} & \multicolumn{2}{c}{tieredImagenet (\%)} \\ \cline{3-6} 
\multirow{-2}{*}{Method} & \multirow{-2}{*}{\begin{tabular}[c]{@{}c@{}}Vision\\ Backbone\end{tabular}} & \multicolumn{1}{c|}{5-way 1-shot} & 5-way 5-shot & \multicolumn{1}{c|}{5-way 1-shot} & 5-way 5-shot \\ \hline
EPNet+SSL & WRN-28-10 & \multicolumn{1}{c|}{79.22 ± 0.92} & 88.05 ± 0.51 & \multicolumn{1}{c|}{83.69 ± 0.99} & 89.34 ± 0.59 \\
EASY & 3×ResNet12 & \multicolumn{1}{c|}{83.02 ± 0.23} & 88.57 ± 0.12 & \multicolumn{1}{c|}{84.29 ± 0.24} & 89.76 ± 0.14 \\
PT+MAP & {\color[HTML]{000000} WRN/DenseNet121} & \multicolumn{1}{c|}{{\color[HTML]{000000} 82.92 ± 0.26}} & {\color[HTML]{000000} 88.82 ± 0.13} & \multicolumn{1}{c|}{{\color[HTML]{000000} 85.67 ± 0.26}} & {\color[HTML]{000000} 90.45 ± 0.14} \\
cluster-FSL & WRN-28-10 & \multicolumn{1}{c|}{85.74 ± 0.76} & 90.18 ± 0.43 & \multicolumn{1}{c|}{82.63 ± 0.79} & 89.16 ± 0.35 \\
HCTransformers & ViT-S & \multicolumn{1}{c|}{74.62 ± 0.20} & 89.19 ± 0.13 & \multicolumn{1}{c|}{79.57 ± 0.20} & 91.72 ± 0.11 \\
PEMnE-BMS* & DenseNet121 & \multicolumn{1}{c|}{85.54} & 91.53 & \multicolumn{1}{c|}{86.07 ± 0.25} & 91.09 ± 0.14 \\
CLIP\_LP+LN & {\color[HTML]{000000} ViT-B/16} & \multicolumn{1}{c|}{{\color[HTML]{000000} 92.08}} & {\color[HTML]{000000} 97.94} & \multicolumn{1}{c|}{{\color[HTML]{000000} -}} & {\color[HTML]{000000} -} \\
P\textgreater{}M\textgreater{}F (with ext. data) & ViT-B/16 & \multicolumn{1}{c|}{95.30} & 98.40 & \multicolumn{1}{c|}{-} & - \\
\textbf{SgVA-CLIP (ours)} & {\color[HTML]{000000} ViT-B/16} & \multicolumn{1}{c|}{{\color[HTML]{000000} \textbf{97.95 ± 0.19}}} & {\color[HTML]{000000} \textbf{98.72 ± 0.13}} & \multicolumn{1}{c|}{{\color[HTML]{000000} \textbf{95.73 ± 0.37}}} & {\color[HTML]{000000} \textbf{96.21 ± 0.37}} \\ \bottomrule[1pt]
\end{tabular}
\label{table1}
\end{table*}

\begin{table*}
\caption{The benefit of knowledge distillation.}
\centering
\setlength{\tabcolsep}{1.60mm}
\renewcommand\arraystretch{1.3}
\begin{tabular}{c|c|c|c|c|c|c|c}
\toprule[1pt]
 &  &  &  &  &  &  &  \\
\multirow{-2}{*}{Method} & \multirow{-2}{*}{$L_{KD}$} & \multirow{-2}{*}{\begin{tabular}[c]{@{}c@{}}Vision\\ Backbone\end{tabular}} & \multirow{-2}{*}{\begin{tabular}[c]{@{}c@{}}ImageNet (\%)\\ 16-shot\end{tabular}} & \multirow{-2}{*}{\begin{tabular}[c]{@{}c@{}}SUN397 (\%)\\ 16-shot\end{tabular}} & \multirow{-2}{*}{\begin{tabular}[c]{@{}c@{}}Vision\\ Backbone\end{tabular}} & \multirow{-2}{*}{\begin{tabular}[c]{@{}c@{}}ImageNet (\%)\\ 16-shot\end{tabular}} & \multirow{-2}{*}{\begin{tabular}[c]{@{}c@{}}SUN397 (\%)\\ 16-shot\end{tabular}} \\ \hline
Vision-specific w/o KD & × &  & 63.61 & {\color[HTML]{343434} 69.00} &  & {\color[HTML]{000000} 50.11} & {\color[HTML]{000000} 61.70} \\
\textbf{Vision-specific w/ KD} & \checkmark &  & \textbf{66.95 (+3.34)} & {\color[HTML]{343434} \textbf{71.22 (+2.22)}} &  & \textbf{58.75 (+8.64)} & \textbf{66.56 (+4.86)} \\ \cline{1-2} \cline{4-5} \cline{7-8} 
SgVA-CLIP w/o KD & × &  & 72.94 & 76.12 &  & 64.42 & 71.24 \\
\textbf{SgVA-CLIP w/ KD} & \checkmark & \multirow{-4}{*}{ViT-B/16} & {\color[HTML]{000000} \textbf{73.30 (+0.36)}} & {\color[HTML]{000000} \textbf{76.42 (+0.30)}} & \multirow{-4}{*}{ResNet50} & {\color[HTML]{000000} \textbf{65.70 (+1.28)}} & {\color[HTML]{000000} \textbf{71.99 (+0.75)}} \\   \bottomrule[1pt]
\end{tabular}
\label{table3}
\end{table*}

\begin{figure*}[h]

\centering
\includegraphics[width=1\linewidth]{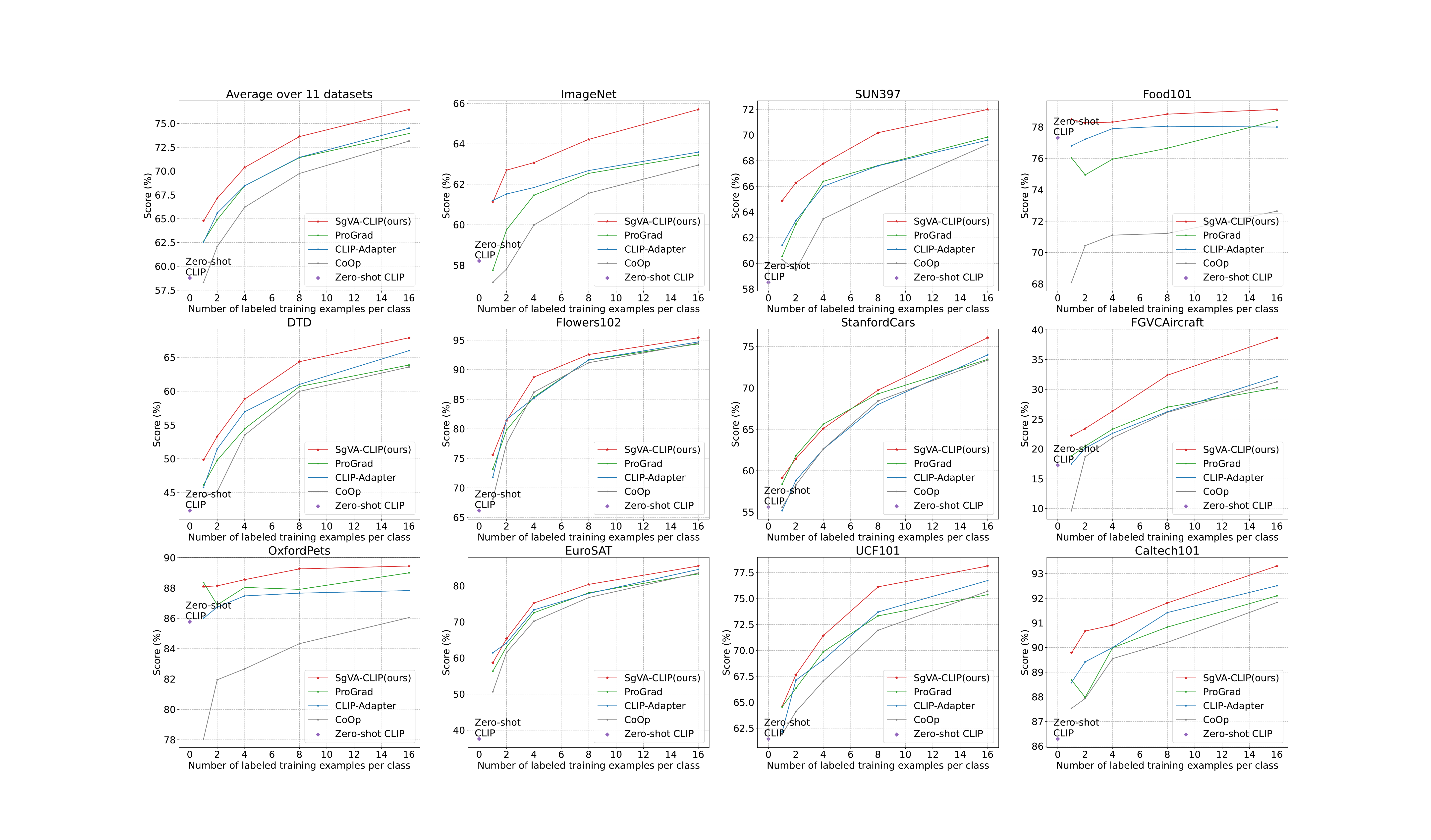}
\caption{Comparison with state-of-the-art few-shot learning methods (without base classes) on 11 datasets.}
\label{Fig4}
\end{figure*}

\section{Experiment and Results}
\subsection{Datasets}
%
For the standard meta-learning scenario, we choose miniImagenet \cite{NIPS2016_90e13578} and tieredImagenet \cite{ren2018meta}, which are common benchmarks for few-shot learning.
The purpose is to validate the generalization ability of our model on novel classes.
The miniImagenet dataset contains 100 classes sampled from ILSVRC-2012 \cite{russakovsky2015imagenet}, and is split to 64, 16, 20 classes for training, validation, and testing respectively.
Similarly, the tieredImagenet dataset includes 608 classes sampled from ILSVRC-2012, and is divided into 351, 97, 160 classes for training, validation, and testing respectively.
The setting of tieredImagenet is more challenging, because the base classes and novel classes come from different super categories. \par

For the scenario without base classes, we follow CLIP \cite{radford2021learning} and CoOp \cite{zhou2022learning} to select 11 image classification datasets to evaluate the performance, namely ImageNet \cite{deng2009imagenet}, StanfordCars \cite{krause20133d}, UCF101 \cite{soomro2012ucf101}, Caltech101 \cite{fei2004learning}, Flowers102 \cite{nilsback2008automated}, SUN397 \cite{xiao2010sun}, DTD \cite{cimpoi2014describing}, EuroSAT \cite{helber2019eurosat}, FGVCAircraft \cite{maji2013fine}, OxfordPets \cite{parkhi2012cats}, and Food101 \cite{bossard2014food}.
These datasets cover a series of diverse visual tasks including the classification of general objects, scenes, actions, and fine-grained categories.

\subsection{Implementation Details}
%
%
For the conventional setting of few-shot learning,
we evaluate our model on two widely used datasets, i.e., miniImagenet and tieredImagenet.
We train the overall framework on base classes for 100 epochs with 5-way 1-shot/5-shot tasks.
And in each episode, 15 query images per class are randomly sampled.
In test phase, we randomly 600 episodes on novel classes, and report the mean accuracy together with the 95\% confidence interval.
For the setting without base classes, 
we follow the few-shot evaluation protocol adopted in CLIP~\cite{radford2021learning} to evaluate the model performance on 11 datasets, and set up 1, 2, 4, 8, 16 shots from all the classes to train the model and then test it on full test set. 
The optimizer is SGD with momentum of 0.9 and weight decay of 0.0005.
%
{
The temperature $\tau_{1}$ is obtained from the pre-trained CLIP, while the temperature $\tau_{2}$ in implicit knowledge distillation is set to 5, whose parameter analysis is shown in Table \ref{table5}. The length of prompt vectors is set to 4, which is same length as the hand-crafted prompt ``a photo of a".}
We conduct all experiments on a single Nvidia V100 GPU.
%
%


%

%
%
%

%
%
%

\subsection{Baselines}
For the miniImagenet and tieredImagenet datasets, we compare our method with 8 baselines, including PEMnE-BMS* \cite{hu2022squeezing}, HCTransformers \cite{he2022attribute}, CLIP\_LP+LN \cite{kim2021adapt}, P\textgreater{M}\textgreater{F} \cite{hu2022pushing}, cluster-FSL \cite{ling2022semi}, PT+MAP \cite{hu2021leveraging}, EPNet \cite{rodriguez2020embedding} and EASY \cite{bendou2022easy} which are state-of-the-art methods on miniImagenet and tieredImagenet.
For the other 11 datasets, we compare our method with 5 baselines, namely Zero-shot CLIP \cite{radford2021learning}, CoOp \cite{zhou2022learning} and CLIP-Adapter \cite{gao2021clip}, {ProGrad \cite{BeierZhu2022PromptalignedGF}}, which are state-of-the-art few-shot learning methods based on VLP models.
For fair comparison, we follow their way of dataset splitting and adopt classification accuracy as the evaluation metric.
%

\subsection{Performance Comparison}
%
Table \ref{table1} shows the results 
of our SgVA-CLIP and other state-of-the-art methods on miniImagenet and tieredImagenet.
%
%
%
Compared to the other methods,
our SgVA-CLIP achieves new state-of-the-art performance on both miniImagenet and tieredImageNet.
Although the P\textgreater{M}\textgreater{F} \cite{hu2022pushing} uses extra data, SgVA-CLIP still outperforms it by 2.65\% and 0.32\% respectively on the 5-way 1-shot and 5-way 5-shot tasks of miniImagenet.
Compared with CLIP\_LP+LN \cite{kim2021adapt}, which uses the same pre-trained parameters of CLIP as ours, SgVA-CLIP exceeds it by 5.87\% and 0.78\% respectively on the miniImagenet.
Besides, on the tieredImageNet, SgVA-CLIP surpasses the second best result by 9.66\% and 4.49\% respectively.
It is worth noting that the improvement of our method is larger when the number of shots is fewer, which demonstrates its generality capability in few-shot learning.\par

%

%

%

%
Fig. \ref{Fig4} shows the comparison results of our method
and 5 recent baselines over 11 datasets with 1, 2, 4, 8, 16 shots.
For fair comparison, we adopt the vision backbone ResNet50 from CLIP as in \cite{zhou2022learning,gao2021clip,zhang2021tip,zhang2021vt}.
Compared with CoOp \cite{zhou2022learning}, our method achieves considerable improvement.
%
CoOp only considers the cross-modal information extraction from a perspective of prompt learning, but neglects the vision-specific information that is more discriminative.
By comprehensively exploiting the vision information and the cross-modal information, our method outperforms CoOp on 11 datasets.
Typically, the average performance gains over all datasets with 1, 2, 4, 8, 16 shots are 6.45\%, 5.07\%, 4.19\%, 3.87\% and 3.32\% respectively. Besides, the average gains over all the shots has reached 7.9\% and 7.09\% respectively on the Food101 and FGVAircraft dataset. \par
%
%

%
{
Compared with ProGrad \cite{BeierZhu2022PromptalignedGF}, which proposes a prompt-aligned gradient updating scheme, our SgVA-CLIP achieves significant performance and exceeds it by 2.21\% on average. Typically, the mean performance gains over 11 datasets with 1, 2, 4, 8, 16 shots are 2.15\%, 2.24\%, 1.93\%, 2.21\% and 2.53\% respectively.
Although a novel gradient updating strategy is proposed from the perspective of overcoming the improperly biased tuning, ProGrad gains limited improvement and cannot capture the more discriminative visual information.
}
Compared with CLIP-Adapter \cite{gao2021clip}, which is a visual adapter based on CLIP, our SgVA-CLIP surpasses it on most datasets.
The average performance gains over CLIP-Adapter across all shots on the 11 datasets is 1.97\%. By training an extra visual bottleneck layer, CLIP-Adapter can enhance the alignment of the visual features with the text features.
However, it only considers the visual information that is related to text.
Different from it, our SgVA-CLIP proposes a semantic-guided adapting mechanism, which produces more discriminative visual features that can well complement the cross-modal features, showing a more promising perspective for few-shot learning.\par
%

%

%
%
%
%
The overall experiment results 
demonstrate that the implicit knowledge distillation can promote the visual adapting and produce more effective task-specific visual features for few-shot learning.
%

\subsection{Ablation Study}
%
%
%
%
%
%
%


\begin{table}[t]
\caption{The significant test of knowledge distillation on five trials.}
\centering
\setlength{\tabcolsep}{0.62mm}
\renewcommand\arraystretch{1.3}
\begin{tabular}{c|c|c|c|c|c|c}
\toprule[1pt]
Method & $L_{KD}$ & \begin{tabular}[c]{@{}c@{}}Vision\\ Bckbone\end{tabular} & \begin{tabular}[c]{@{}c@{}}Mean acc on\\ ImageNet(\%)\\ 16-shot\end{tabular} & std & T value & P value \\ \hline
SgVA-CLIP w/o KD & × & \multirow{2}{*}{ViT-B/16} & 72.94 & 0.36 & \multirow{2}{*}{3.70}& \multirow{2}{*}{3.6e-4} \\ \cline{1-2} \cline{4-5}
\textbf{SgVA-CLIP w/ KD} & \checkmark &  & \textbf{73.30} & 0.11 &  &  \\ \bottomrule[1pt]
\end{tabular}
\label{significant test}
\end{table}

\textbf{Ablation study of the implicit knowledge distillation.}
%
To evaluate the effectiveness of the adopted knowledge distillation, we conduct experiment on ImageNet and SUN397.
%
%
The results are shown in Table \ref{table3}.
We can observe that with ResNet50 as the vision backbone, KD improves the accuracy of vision-based model by 8.64\% on ImageNet and 4.86\% on SUN397.
And KD enhances the SgVA-CLIP by 1.28\% and 0.75\% respectively. 
%
%
We also note that when ResNet50 is used as the vision backbone, the performance gain of KD is greater than using the ViT-B/16 backbone. 

{To do the significance test for knowledge distillation, we compute the P-value by repeating the experiments 5 times with different random seeds. With the backbone of ViT-B/16, the results of five trials on ImageNet (16-shot) are shown in Table \ref{significant test}, where std refers to the standard deviation. The P value of T test was calculated as $3.6e-4<0.05$, demonstrating that there are significant differences between SgVA-CLIP w/ KD and SgVA-CLIP w/o KD.}






\begin{table}[t]
\centering
\caption{Ablation study of direct and implicit knowledge distillation.}
\renewcommand\arraystretch{1.3}
\setlength{\tabcolsep}{1.65mm}
\centering
\begin{tabular}{c|c|c|c|c} 
\toprule[1pt]
\begin{tabular}[c]{@{}c@{}}Knowledge\\ Distillation\end{tabular} & \begin{tabular}[c]{@{}c@{}}Vision\\ Backbone\end{tabular} & \begin{tabular}[c]{@{}c@{}}ImageNet (\%)\\ 16-shot\end{tabular} & \begin{tabular}[c]{@{}c@{}}SUN397 (\%)\\ 16-shot\end{tabular} & Average (\%) \\ 
\hline
Direct & \multirow{2}{*}{ViT-B/16} & 68.98 & 75.70 & 72.34 \\
Implicit &  & \textbf{73.30} & \textbf{76.42} & \textbf{74.86} \\ 
\hline
Direct & \multirow{2}{*}{ResNet50} & 59.88 & 66.38 & 63.13 \\
\textbf{Implicit} &  & \textbf{65.64} & \textbf{71.99} & \textbf{68.82} \\
\bottomrule[1pt]
\end{tabular}
\label{table4}
\end{table}

{
\textbf{Implicit knowledge distillation vs. direct distillation.}
The direct distillation uses the Kullback-Leibler divergence to 
match the sample relations in the cross-modal space 
and the sample relations in the vision space.
As shown in Table \ref{table4}, compared with direct distillation, the proposed implicit knowledge distillation has an average performance gain of {2.52\% and 5.69\% respectively on ViT-B/16 and ResNet50 backbone}.
Because in the direct distillation,
the distribution gap between the two spaces may have a negative impact on the distillation.}
%

\begin{table}[t]
\centering
\caption{The complementarity between the vision space and the cross-modal space.}
\renewcommand\arraystretch{1.3}
\setlength{\tabcolsep}{0.45pt}
\begin{tabular}{c|c|c|c|c|c|c}
\toprule[1pt]
\multirow{2}{*}{Method} & \multirow{2}{*}{\begin{tabular}[c]{@{}c@{}}Vision \\ Backbone\end{tabular}} & \multirow{2}{*}{$L_{cl\_i2t}$} & \multirow{2}{*}{$L_{cl\_i2i}$} & \multicolumn{1}{l|}{\multirow{2}{*}{$L_{KD}$}} & ImageNet & SUN397 \\ \cline{6-7} 
 &  &  &  & \multicolumn{1}{l|}{} & 16-shot & 16-shot \\ \hline
Vision-specific Prediction & ViT-B/16 & × & \checkmark & × & 63.61 & 69.00 \\
Cross-modal Prediction & ViT-B/16 & \checkmark & × & × & 71.48 & 73.50 \\
\textbf{Fused Prediction} & ViT-B/16 & \checkmark & \checkmark & × & \textbf{72.94} & \textbf{76.12} \\ \bottomrule[1pt]
\end{tabular}
\label{table2}
\end{table}

\textbf{Complementarity between the vision space and the cross-modal space.}
%
%
We compare the classification results on ImageNet and SUN397 obtained using the visual features and/or the cross-modal features.
%
%
The results are shown in Table \ref{table2}, which demonstrate that 
comprehensively considering the visual feature and the cross-modal feature is better than simply using the one of them.
%
%
%

%
For the 16-shot learning task, fusing the results predicted from the vision features and the cross-modal features will increase the accuracy by 1.46\% and 2.62\% on ImageNet and SUN397 when compared with the cross-modal results, and the accuracy is increased by 9.33\% and 7.12\% respectively when compared with the vision-based results.\par

{
\textbf{Ablation study of the visual adapting layer and the learnable prompt.}
With the visual adapter removed, the discriminative adapted visual features are replaced by the pre-trained visual features. And when prompt is not learnable, we follow CLIP \cite{radford2021learning} and use hand-crafted prompt, i.e. 'a photo of a'. Note that the learnable continuous prompt has the same length as the hand-crafted prompt, i.e. 4. And the 5-way 1-shot and 5-way 5-shot tasks are abbreviated as 5w-1s and 5w-5s respectively in Table \ref{ablation}.\par
The baseline is that of removing both the Visual Adapter Layer and the learnable prompt. As shown in Table \ref{ablation}, the Visual Adapter Layer improves the accuracy on the 5-way 1-shot task of miniImagenet from 93.07\% to 95.71\%, and the learnable prompt elevates the accuracy to 96.63\%. With the Visual Adapter Layer and the learnable prompt applied together, the accuracy rate reaches 97.95\%, 4.88\% higher than the baseline.}

\begin{table}[t]
\centering
\caption{Ablation study of Adapter and Prompt.}
\renewcommand\arraystretch{1.1}
\setlength{\tabcolsep}{1.25mm}
\begin{tabular}{c|c|c|cc|cc}
\toprule[1pt]
 &  &  & \multicolumn{2}{c|}{miniImagenet (\%)} & \multicolumn{2}{c}{tieredImagenet (\%)} \\ \cline{4-7} 
\multirow{-2}{*}{\begin{tabular}[c]{@{}c@{}}Vision\\ Backbone\end{tabular}} & \multirow{-2}{*}{\begin{tabular}[c]{@{}c@{}}Visual\\ Adapter\end{tabular}} & \multirow{-2}{*}{\begin{tabular}[c]{@{}c@{}}Learnable\\ Prompt\end{tabular}} & \multicolumn{1}{c|}{5w-1s} & 5w-5s & \multicolumn{1}{c|}{5w-1s} & 5w-5s \\ \hline
 & × & × & \multicolumn{1}{c|}{93.07} & 97.45 & \multicolumn{1}{c|}{89.72} & 93.73 \\
 & \checkmark & × & \multicolumn{1}{c|}{95.71} & 97.82 & \multicolumn{1}{c|}{92.30} & 95.25 \\
 & × & \checkmark & \multicolumn{1}{c|}{96.63} & 97.80 & \multicolumn{1}{c|}{94.88} & 95.21 \\
\multirow{-4}{*}{ViT-B/16} & \textbf{\checkmark} & \textbf{\checkmark} & \multicolumn{1}{c|}{{\color[HTML]{000000} \textbf{97.95}}} & {\color[HTML]{000000} \textbf{98.72}} & \multicolumn{1}{c|}{{\color[HTML]{000000} \textbf{95.73}}} & {\color[HTML]{000000} \textbf{96.21}} \\ \bottomrule[1pt]
\end{tabular}
\label{ablation}
\end{table}


\begin{table}[t]
\centering
\caption{Comparison of different vision backbones.}
\renewcommand\arraystretch{1.1}
\setlength{\tabcolsep}{1.5mm}
\begin{tabular}{c|c|c|c|c} 
\toprule[1pt]
\multirow{2}{*}{\begin{tabular}[c]{@{}c@{}}Vision\\Backbone\end{tabular}} & \multirow{2}{*}{Method} & ImageNet (\%) & SUN397 (\%) & \multirow{2}{*}{Average (\%)} \\ 
\cline{3-4}
 &  & 16-shot & 16-shot &  \\ 
\hline
\multirow{2}{*}{ResNet50} & CoOp & 62.95 & 69.26 & 66.11 \\
 & \textbf{SgVA-CLIP} & \textbf{65.70} & \textbf{71.99} & \textbf{68.85} \\ 
\hline
\multirow{2}{*}{ResNet101} & CoOp & 66.60 & 71.19 & 68.90 \\
 & \textbf{SgVA-CLIP} & \textbf{68.51} & \textbf{73.00} & \textbf{70.76} \\ 
\hline
\multirow{2}{*}{ViT-B/32} & CoOp & 66.85 & 72.38 & 69.62 \\
 & \textbf{SgVA-CLIP} & \textbf{68.26} & \textbf{74.04} & \textbf{71.15} \\ 
\hline
\multirow{2}{*}{ViT-B/16} & CoOp & 71.92 & 75.29 & 73.61 \\
 & \textbf{SgVA-CLIP} & \textbf{73.30} & \textbf{76.42} & \textbf{74.86} \\
\bottomrule[1pt]
\end{tabular}
\label{table7}
\end{table}

\begin{figure}[H]
\centering
\includegraphics[width=0.99\linewidth]{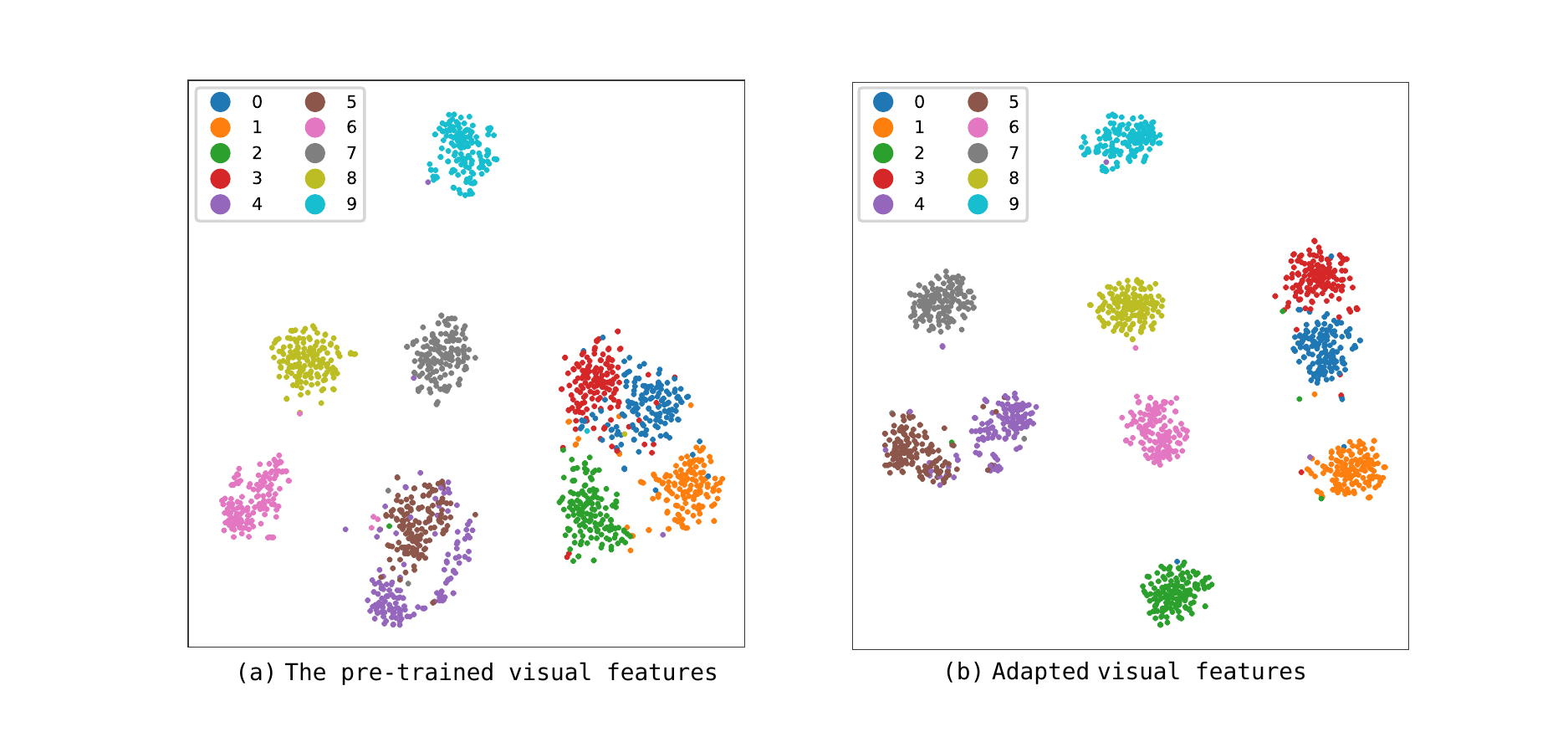}
\caption{Visualization of the distribution of the pre-trained visual features and adapted visual features.}
\label{fig5}
\end{figure}

{
\subsection{Results on Different Vision Backbones}
The results in Fig. 4 are based on the backbone ResNet50 for fair comparison with other methods, but SgVA-CLIP is also effective on other vision backbones.
Considering that only CoOp \cite{zhou2022learning} did a comprehensive analysis experiment of ViT-B/16, ViT-B/32, ResNet50 and ResNet101 backbones, we report more comparison results with CoOp as shown in Table \ref{table7}.
SgVA-CLIP surpasses CoOp by {2.74\%, 1.86\%, 1.53\% and 1.25\%} on average respectively on ViT-B/16, ViT-B/32, ResNet50 and ResNet101 backbones.

\subsection{Parameter Analysis}
%
%
\begin{table}[t]
\centering
\caption{Effect of temperature in knowledge distillation.}
\renewcommand\arraystretch{1.1}
\setlength{\tabcolsep}{1.15mm}
\begin{tabular}{c|c|c|c|c} 
\toprule[1pt]
\begin{tabular}[c]{@{}c@{}}Distillation\\ Temperature $\tau_{2}$\end{tabular} & \begin{tabular}[c]{@{}c@{}}Vision\\ Backbone\end{tabular} & \begin{tabular}[c]{@{}c@{}}ImageNet (\%)\\ 16-shot\end{tabular} & \begin{tabular}[c]{@{}c@{}}SUN397 (\%)\\ 16-shot\end{tabular} & Average (\%) \\ 
\hline
5 & \multirow{5}{*}{ResNet50} & 65.64 & \textbf{71.99} & \textbf{68.82} \\
10 &  & \textbf{65.70} & 71.86 & 68.78 \\
15 &  & 65.28 & 71.58 & 68.43 \\
20 &  & 64.78 & 71.49 & 68.14 \\
25 &  & 64.63 & 71.37 & 68.00 \\ \bottomrule[1pt]

\end{tabular}
\label{table5}
\end{table}

\textbf{Parameter analysis of the temperature $\tau_{2}$ in distillation.} To analyze the effect of the temperature $\tau_{2}$ in distillation,
we conduct experiments on ImageNet and SUN397 with different settings.
It is worth noting that the tenperature $\tau_{1}$ used in the cross-modal contrastive loss and vision-specific contrastive loss is outside the scope of parameter analysis because it is a pre-trained parameter of CLIP.
The temperature $\tau_{2}$ controls the smoothness of the soft labels.
As shown in Table \ref{table5}, we obtain the best performance when $\tau_{2}$ is 5.
%
%

%
\textbf{Parameter analysis of the hidden dimension in the visual adapting layer.} Table~\ref{table6} shows the results of using different {hidden dimensions of the visual adapter layer}.
%
%
%
%
%
We observe that either too small or too large 
dimension will deteriorate the performance and the best adapter dimension is 4096, which is able to preserve enough visual information without redundancy.
Therefore, we set the dimension to 4096 in the experiment.

\begin{table}[t]
\renewcommand\arraystretch{1.05}
\centering
\caption{Effect of the hidden dimension in the
visual adapting layer.}
\setlength{\tabcolsep}{1.65mm}
\begin{tabular}{l|c|c|c|c}
\toprule[1pt]
\multirow{2}{*}{\begin{tabular}[c]{@{}c@{}}Vision\\  Backbone\end{tabular}} & \multirow{2}{*}{\begin{tabular}[c]{@{}c@{}}Hidden\\ Dimension\end{tabular}} & ImageNet (\%) & SUN397 (\%) & \multirow{2}{*}{Average (\%)} \\ \cline{3-4}
 &  & 16-shot & 16-shot &  \\ \hline
\multirow{5}{*}{ResNet50} & 512 & 65.28 & 71.19 &  68.24\\
& 1024 & 65.16 & 71.37 & 68.27 \\
 & 2048 & 65.53 & 71.86 & 68.70 \\
 & \textbf{4096} & \textbf{65.64} & \textbf{71.99} & \textbf{68.82} \\
 & 8192 & 65.48 & 71.96 & 68.72 \\ \bottomrule[1pt]
\end{tabular}
\label{table6}
\end{table}

\subsection{Visualization}
{In Fig. \ref{fig5}, we sample 10 classes} and display the distribution of the adapted visual features obtained by SgVA-CLIP and the pre-trained visual features obtained by CLIP. The visualization results show that the adapted visual features are more discriminative than the pre-trained visual features, and thus it is important to consider the adapted visual features in few-shot classification.
}


\section{Conclusion}
We present SgVA-CLIP, a new VLP-based few-shot classification approach,
which can comprehensively consider unimodal vision correlation and cross-modal image-text correlation.
%
%
SgVA-CLIP focuses on the contrastive learning in two spaces and knowledge distillation between them, so that fine-grained cross-modal knowledge sharing can promote the learning of discriminative adapted unimodal vision representations.
With the CLIP model frozen and only a few external parameters updated, the representation ability of CLIP can be quickly migrated to downstream classification tasks by a few labeled data.
According to the experimental results, SgVA-CLIP outperforms competitive baselines on 13 datasets under different few-shot settings.
%
%
In future work, we will combine SgVA-CLIP with other efficient tuning methods and explore the application of SgVA-CLIP in more downstream tasks.
\ifCLASSOPTIONcaptionsoff
  \newpage
\fi



%
\bibliographystyle{ieeetr}
\bibliography{ref}

\end{document}